\documentclass[10pt,twocolumn,letterpaper]{article}

\usepackage[margin=0.75in,top=0.8in,bottom=0.9in]{geometry}
\usepackage[T1]{fontenc}
\usepackage{lmodern}
\usepackage{microtype}
\usepackage{graphicx}
\usepackage{booktabs}
\usepackage{multirow}
\usepackage{amsmath,amssymb}
\usepackage{xcolor}
\usepackage{subcaption}
\usepackage[numbers,sort&compress]{natbib}
\usepackage{url}
\usepackage[colorlinks=true,linkcolor=blue!60!black,citecolor=blue!60!black,urlcolor=blue!60!black]{hyperref}
\usepackage{enumitem}
\usepackage{placeins}
\usepackage{tikz}
\usetikzlibrary{arrows.meta,positioning,calc,fit,backgrounds,decorations.pathreplacing}
\setlist{nosep,leftmargin=*}
\usepackage{caption}
\graphicspath{{figures/}}

\newcommand{\ours}{\textsc{LiveMirror}}
\newcommand{\akt}{asynchronous keyframe try-on}
\newcommand{\AKT}{Asynchronous keyframe try-on}
\newcommand{\longmask}{\texttt{long}}
\newcommand{\uppermask}{\texttt{upper}}
\newcommand{\overallmask}{\texttt{overall}}

\title{\textbf{\ours{}: Real-Time Diffusion Virtual Try-On on a Consumer GPU\\and Its Self-Supervised Adaptation to Indian Ethnic Wear}}

\author{%
Jigyansha Jeevan Rout\\
School of Electronics Engineering\\
KIIT Deemed to be University, Bhubaneswar, India\\
\texttt{jigyanshajeevan@gmail.com}
\and
Sukanta Kumar Sabut\\
School of Electronics Engineering\\
KIIT Deemed to be University, Bhubaneswar, India\\
\texttt{sukanta.sabatfet@kiit.ac.in}%
}
\date{}

\begin{document}
\maketitle

\begin{abstract}
Diffusion virtual try-on (VTO) models produce convincing garment transfers from a single product photograph, but they take hundreds of milliseconds to seconds per image, so every deployment today is either offline (upload a photo, wait) or falls back to augmented-reality mirrors that overlay rigged 3D garments and therefore need a 3D asset per product. We present \ours{}, a live-camera try-on mirror that renders a flat garment photograph onto a moving person with a diffusion model on one 16\,GB consumer GPU. Its core is \emph{\akt{}}: generation never blocks the display, every displayed frame is an optical-flow warp of the latest generated keyframe, and a keyframe is only ever replaced through a cross-fade. With this design the display runs at camera rate (25\,FPS at 720p) while a new try-on arrives every 370--490\,ms; we measure the residual flicker, give a stage-by-stage latency budget, and report which inference accelerations (step count, consistency LoRA, tiny decoder, compilation) keep garment prints intact. We then show that the same GPU is enough to fix the model's biggest cultural blind spot. Try-on models trained on VITON-HD and DressCode render a knee-length \emph{kurta} or \emph{sherwani} as a torso-length top or as a dress over bare legs, and no inference-time setting repairs it. A thigh-covering \longmask{} mask, self-supervised pairs built automatically from 24.9k on-model e-commerce photographs, and 3.2 GPU-hours of attention-only fine-tuning of CatVTON halve the perceptual error on ethnic wear (LPIPS 0.137 to 0.077 on IndoFashion, 0.135 to 0.073 on Atlas, 200 held-out pairs per source) with no regression on Western tops. A two-pass \emph{hem-aware} mask, which cuts the thigh mask just below the hem found in a first pass, then removes the remaining failure of these models, repainting the trousers between hem and knee in another colour, on all but 3--4\% of pairs at no training cost; in the mirror the hem comes from the previous keyframe for free. The adapted model also beats FASHN VTON 1.5, a recent open-weights try-on model that truncates these garments to hip length, on every aggregate ethnic-wear metric at about $20\times$ lower inference cost, and its weights drop straight into the mirror. Code for the pipeline, pair construction, and training may be made available on request.
\end{abstract}

\section{Introduction}
\label{sec:intro}

Image-based virtual try-on renders a reference garment onto a photograph of a person while preserving identity, pose, and the untouched parts of the scene. Latent-diffusion inpainting models~\cite{zhu2023tryondiffusion,kim2024stableviton,xu2024ootdiffusion,choi2024idmvton,chong2024catvton} now do this convincingly from a single flat product photograph. What they do not do is run live. Even the lightest of them needs several hundred milliseconds per frame on a consumer GPU, and a person in front of a mirror moves in every one of those frames. Products that call themselves ``virtual mirrors'' therefore either run the model offline on an uploaded photo or overlay a rigged 3D garment with an augmented-reality tracker, which requires a 3D asset to be authored for every product and cannot show fabric, print, or drape the way a generative model can.

\ours{} closes that gap. It is a live-camera try-on mirror in which the garment is a flat photograph, the renderer is a diffusion model, and the whole system, tracker, parsers, and model, runs on one 16\,GB GPU. The design observation is simple: the display loop and the generator should not share a clock. We call the resulting scheme \emph{\akt{}} (Section~\ref{sec:akt}): the diffusion model generates try-on \emph{keyframes} asynchronously and never blocks the display; every displayed frame is the latest keyframe warped onto the current camera frame with dense optical flow; and a new keyframe only enters through a short cross-fade. Generation latency then sets how fresh the garment is, not how smooth the mirror is. On an RTX~4080 the mirror displays at 25\,FPS at 720p while a keyframe arrives every 370--490\,ms, and the garment stays attached to the body within about two grey levels of frame-to-frame change between keyframes (Section~\ref{sec:livemeas}).

\begin{figure}[t]
\centering
\includegraphics[width=\linewidth]{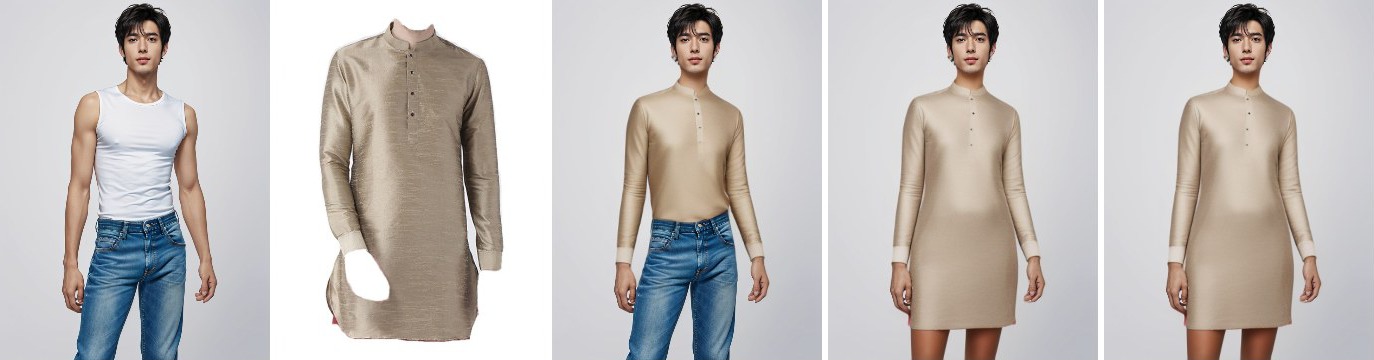}
\caption{\textbf{Why inference-time fixes fail on ethnic wear.} Person, kurta, and CatVTON (VITON-HD weights) outputs with the \uppermask{} mask (kurta becomes a shirt), the \overallmask{} mask (dress prior: bare legs), and our \longmask{} mask without fine-tuning (correct hem, but the hidden trouser band becomes skin). The base model has no concept of a long garment over trousers.}
\label{fig:motivation}
\end{figure}

A mirror that only knows Western clothing is of limited use in the market where we deployed it. The strongest try-on models are trained on VITON-HD~\cite{choi2021vitonhd} and DressCode~\cite{morelli2022dresscode}, whose garment taxonomy is tops, bottoms, and one-piece dresses. South Asian ethnic wear (kurta, sherwani, pathani, anarkali, kurti), worn by hundreds of millions of people, is a \emph{long upper garment worn over visible trousers}, and this silhouette does not exist in those datasets. Figure~\ref{fig:motivation} shows the consequence: with the standard upper-body mask the kurta is truncated to a shirt; with a full-body mask the model's dress prior produces bare legs; with an intermediate mask the hidden thigh band is filled with skin. The failure is in the learned prior, so it cannot be fixed by any inference-time setting. Fixing it by training is blocked by data, since no paired flat-lay/on-model dataset exists for these garments, while unpaired on-model e-commerce photographs are plentiful~\cite{rajput2021indofashion}.

We show that the same consumer GPU is enough to close this gap too. Our contributions are:
\begin{itemize}
\item \textbf{\ours{}}, a complete live diffusion try-on system on one 16\,GB GPU, built around \emph{\akt{}}, a named and defined display scheme (three rules, Section~\ref{sec:akt}) that decouples display rate from generation latency. We measure its flicker against alternatives, give a stage-by-stage latency budget, and report which accelerations keep garment prints intact and which destroy them. To our knowledge this is the first measured, open description of a diffusion-based live try-on mirror on consumer hardware.
\item A \longmask{} clothing-agnostic mask (torso, arms, and DensePose thighs; lower legs kept) that lets an inpainting try-on model express ``long garment over trousers'' at all, with a per-garment rule for selecting it, and a two-pass \emph{hem-aware} variant that keeps the trousers below the hem visible and cuts the trouser-recolouring failure rate from 16--17\% to 3--4\% of pairs.
\item A fully automatic pair-construction pipeline that turns an on-model photograph into a self-supervised training triplet (parser cut-out as garment card, the photograph as target, quality filter), yielding 24.9k ethnic pairs from two public sources with no annotation.
\item A two-stage attention-only fine-tune of CatVTON~\cite{chong2024catvton} on these pairs mixed with VITON-HD, 3.2 GPU-hours on the same card, that roughly halves perceptual error on ethnic wear with no regression on Western tops, outperforms the open-weights FASHN VTON 1.5 on ethnic wear at $20\times$ lower cost, and whose weights drop into the mirror unchanged.
\end{itemize}

We also report what did not work and what is unmeasured: the hidden trouser colour under the thigh band remains a guess, the flicker study is on a synthetic video, and the held-out evaluation is small (Section~\ref{sec:limitations}).

\section{Related Work}
\label{sec:related}

\paragraph{Image-based virtual try-on.}
Early work warped the garment to the body with thin-plate splines and composited it with a generator~\cite{han2018viton,wang2018cpvton,choi2021vitonhd}. Diffusion models replaced explicit warping with learned inpainting: TryOnDiffusion~\cite{zhu2023tryondiffusion} uses a parallel UNet, StableVITON~\cite{kim2024stableviton} and OOTDiffusion~\cite{xu2024ootdiffusion} inject garment features through cross-attention, IDM-VTON~\cite{choi2024idmvton} adds a full reference UNet and IP-Adapter conditioning on SDXL, and CatVTON~\cite{chong2024catvton} shows that concatenating the garment latent next to the person latent along the spatial axis, and training only the self-attention layers (49.6M parameters), matches heavier designs. We build on CatVTON because it is the lightest at inference and its trainable set fits a consumer GPU. All of these models train on VITON-HD~\cite{choi2021vitonhd} and DressCode~\cite{morelli2022dresscode} and inherit their garment taxonomy.

\paragraph{Real-time and video diffusion.}
StreamDiffusion~\cite{kodaira2023streamdiffusion} pipelines denoising steps across a batch of incoming frames and reaches interactive rates for image-to-image on a datacentre GPU, but every output is a fresh sample and it has no notion of a garment or a locked person. Video try-on models~\cite{fang2024vivid,chong2025catv2ton} add temporal attention to produce coherent clips and are strictly offline: they consume a recorded video and run seconds per clip on 24--80\,GB cards. Streaming video generators with KV caching~\cite{huang2025selfforcing} are the likely long-term path to live generation, but the smallest open model needs about 23\,GB of weights and a 24\,GB GPU. Commercial ``smart mirrors'' instead track the body and overlay a rigged 3D garment; they are real-time but need a 3D asset per product and cannot reproduce print or fabric from a photograph. \ours{} sits in the gap: a flat photograph in, a live mirror out, on a 16\,GB card, by keeping an off-the-shelf image try-on model and solving the latency on the display side.

\paragraph{Datasets and cultural coverage.}
VITON-HD has 13.7k pairs of women's upper-body tops; DressCode has 53.8k pairs in three categories. IndoFashion~\cite{rajput2021indofashion} is a 106k-image classification dataset of Indian ethnic wear with 15 categories, including kurta, sherwani, Nehru jacket, and women's kurta, but has no garment/person pairs. Marqo Atlas~\cite{marqo2024atlas} is a multilingual e-commerce dataset with a subset of Indian ethnic listings. We use both as sources of on-model imagery. To our knowledge no prior try-on work targets long-over-trousers garments or evaluates on South Asian ethnic wear.

\paragraph{Self-supervised pairs and parsing.}
Using the same photograph as source and target, with the garment region masked, is the standard training signal in inpainting try-on~\cite{choi2021vitonhd,chong2024catvton}; the garment side of the pair is normally a separately shot flat-lay. We replace the flat-lay with a human-parser cut-out of the garment from the same image, following the ``model-worn garment'' conditioning that several models accept at inference time~\cite{choi2024idmvton,fashn2025vton}. We use SCHP~\cite{li2020schp} for parsing and DensePose~\cite{guler2018densepose} for the thigh region of the mask, as CatVTON's own masker does.

\paragraph{Fast inference.}
Latent consistency models and LCM-LoRA~\cite{luo2023lcm,luo2023lcmlora} reduce sampling to 4--8 steps; TAESD~\cite{taesd} replaces the VAE decoder with a tiny one. We use both and report their effect on garment prints (Section~\ref{sec:livemeas}).

\section{\ours{}: a live diffusion try-on mirror}
\label{sec:system}

\subsection{\AKT{}}
\label{sec:akt}
A diffusion try-on model at $384\times512$ takes 10--15 camera frames to produce one result on an RTX~4080. Pasting each result into the frame that is current when it arrives makes the garment lag and jump, and running the model synchronously drops the mirror to 2--3\,FPS. \AKT{} resolves this with three rules:
\begin{enumerate}
\item \textbf{Generation never blocks display.} The diffusion model runs in its own thread on a person crop and produces \emph{keyframes}; the display loop runs on every camera frame and never waits for it.
\item \textbf{Every displayed frame is a warp of the latest keyframe.} The display loop computes dense optical flow from the keyframe's source frame to the current frame and warps the keyframe's try-on patch forward, so the garment follows the body between keyframes.
\item \textbf{A keyframe is only replaced through a cross-fade.} When a new keyframe arrives it is blended in over a fixed number of frames; the diffusion seed is fixed per tracked person so that consecutive keyframes differ as little as the input allows.
\end{enumerate}
\begin{figure}[t]
\centering
\begin{tikzpicture}[x=0.056mm, y=1mm, font=\scriptsize,
  tick/.style={line width=0.4pt},
  gbar/.style={draw=blue!50!black, fill=blue!12, rounded corners=0.6pt, line width=0.5pt},
  arr/.style={-{Stealth[length=1.3mm]}, line width=0.5pt},
  lbl/.style={inner sep=1pt}]
\draw[-{Stealth[length=1.5mm]}, black!70] (0,0) -- (1080,0) node[right, lbl] {$t$ (ms)};
\foreach \t in {0,200,...,1000} {\draw[black!70] (\t,0) -- (\t,-1.2) node[below, lbl, text=black!70, font=\tiny] {\t};}
\fill[orange!18] (0,6) rectangle (400,12);
\fill[orange!40] (400,6) rectangle (600,12);
\fill[orange!18] (600,6) rectangle (820,12);
\fill[orange!40] (820,6) rectangle (1020,12);
\fill[orange!18] (1020,6) rectangle (1080,12);
\foreach \i in {0,...,32} {\draw[tick, orange!70!black] (\i*33.3,6) -- (\i*33.3,12);}
\node[anchor=east, lbl, text=orange!70!black, align=right, font=\tiny] at (-15,9) {display\\33\,ms/frame};
\node[lbl, below=0.2mm, text=orange!70!black, font=\tiny] at (200,6) {warp of $K_0$};
\node[lbl, below=0.2mm, text=orange!70!black, font=\tiny] at (500,6) {fade};
\node[lbl, below=0.2mm, text=orange!70!black, font=\tiny] at (710,6) {warp of $K_1$};
\node[lbl, below=0.2mm, text=orange!70!black, font=\tiny] at (920,6) {fade};
\node[anchor=east, lbl, text=blue!50!black, align=right, font=\tiny] at (-15,25) {generator\\async.};
\draw[gbar] (0,22) rectangle (400,28);    \node[lbl, text=blue!50!black] at (200,25) {$K_1 \leftarrow f_0$};
\draw[gbar] (400,22) rectangle (820,28);  \node[lbl, text=blue!50!black] at (610,25) {$K_2 \leftarrow f_{12}$};
\draw[gbar] (820,22) rectangle (1080,28); \node[lbl, text=blue!50!black] at (950,25) {$K_3 \leftarrow f_{25}$};
\foreach \x in {0,400,820} {\draw[arr, blue!50!black] (\x,12) -- (\x,22);}
\foreach \x in {400,820} {\draw[arr, orange!70!black] ([xshift=1.2mm]\x,22) -- ([xshift=1.2mm]\x,12);}
\node[lbl, text=blue!50!black, anchor=west, font=\tiny] at (8,17.5) {latest frame in};
\node[lbl, text=orange!70!black, anchor=west, font=\tiny] at (430,17.5) {$K_1$ out $\to$ fade};
\node[lbl, text=orange!70!black, anchor=west, font=\tiny] at (850,17.5) {$K_2$ out};
\draw[decorate, decoration={brace, amplitude=1mm, mirror}, black!60] (0,-5.5) -- (600,-5.5);
\node[lbl, text=black!60, font=\tiny, below=1.2mm] at (300,-5.5) {age of the garment shown at $t{=}600$: generation + fade};
\end{tikzpicture}
\caption{\textbf{\akt{} on a time axis.} Orange: the display loop emits a frame every 33\,ms; each is an optical-flow warp of the latest keyframe, and a new keyframe enters only through a six-frame cross-fade (darker). Blue: the generator takes the latest frame whenever it becomes free and runs for 370--490\,ms without ever blocking the display. Generation latency sets the \emph{age} of the garment on screen, not the frame rate.}
\label{fig:timing}
\end{figure}
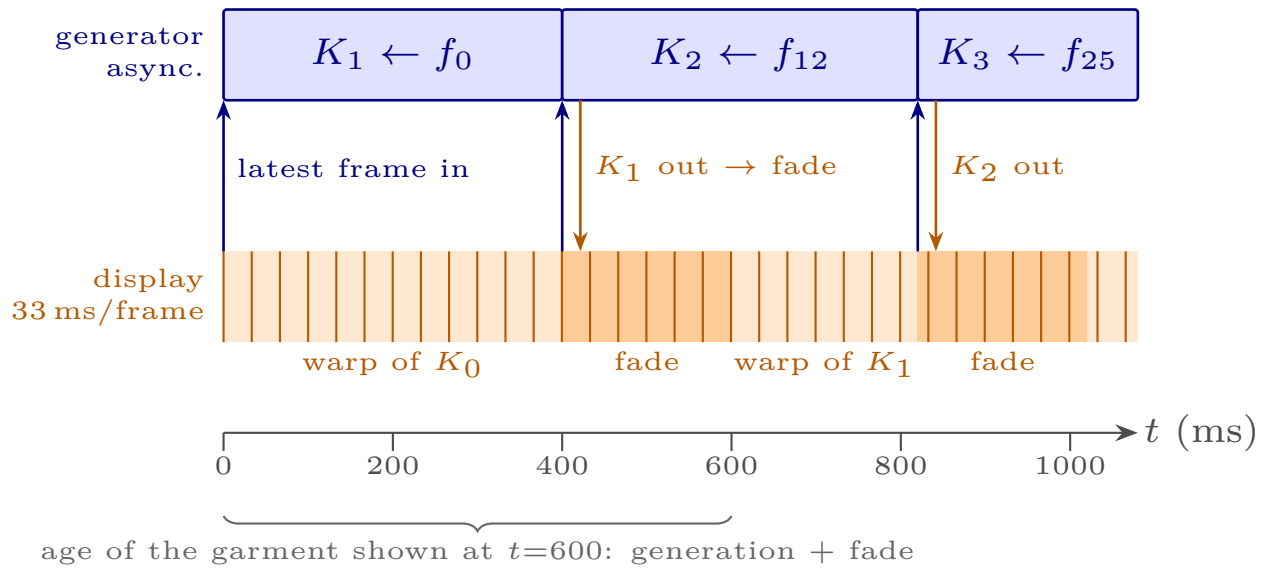

Figure~\ref{fig:timing} draws the three rules on a time axis. Under them the display rate is the camera rate, and generation latency sets only the \emph{age} of the garment on screen. Generation can therefore be as slow as the quality setting demands, and can be throttled: in \emph{hold} mode a keyframe is kept for a few seconds and regenerated only when the garment changes, the person moves more than a quarter of the crop width, or the user asks, which reduces GPU load to one generation every few seconds at no visible cost.

\subsection{Pipeline}
\label{sec:pipeline}
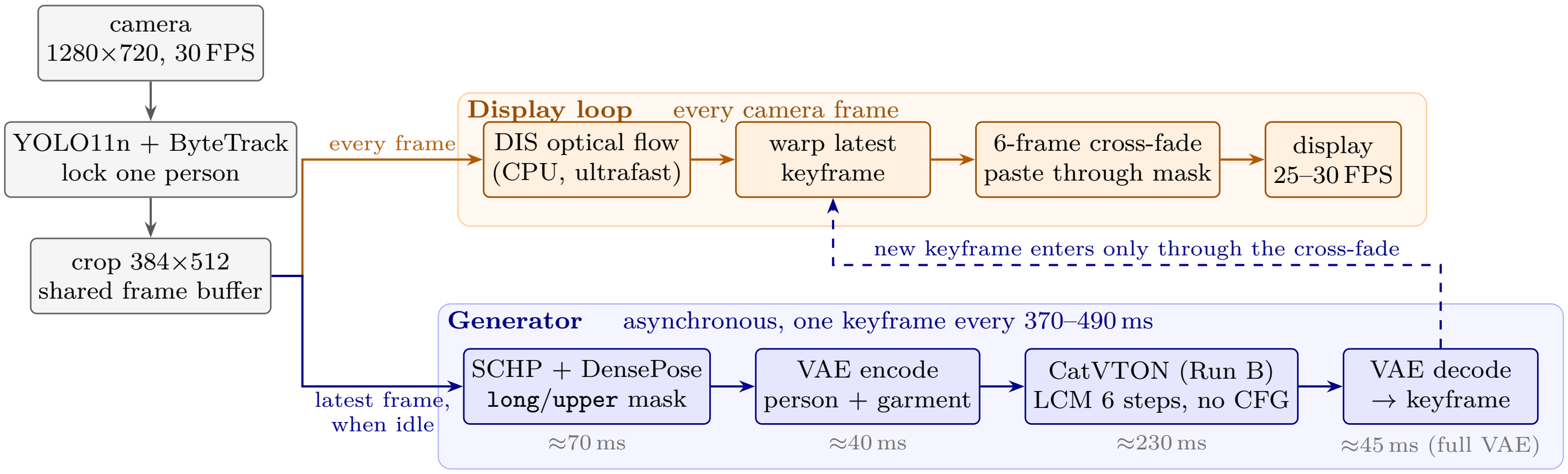
\begin{figure*}[t]
\centering
\begin{tikzpicture}[
  font=\footnotesize, node distance=4.5mm and 5mm,
  box/.style={draw=black!60, line width=0.5pt, rounded corners=2pt, minimum height=8.5mm, minimum width=22mm, inner sep=2.5pt, align=center, fill=white},
  capb/.style={box, fill=black!4},
  disp/.style={box, fill=orange!12, draw=orange!60!black},
  gen/.style={box, fill=blue!10, draw=blue!50!black},
  ms/.style={font=\scriptsize, text=black!55, inner sep=1pt},
  lane/.style={rounded corners=4pt, inner xsep=2.8mm, inner ysep=3mm},
  lanelabel/.style={anchor=north west, font=\footnotesize\bfseries, inner sep=0pt},
  arr/.style={-{Stealth[length=1.8mm]}, line width=0.7pt, black!65},
  darr/.style={arr, orange!70!black},
  garr/.style={arr, blue!55!black},
  lbl/.style={font=\scriptsize, inner sep=1pt, align=center}
]
\node[capb] (cam)  {camera\\$1280{\times}720$, 30\,FPS};
\node[capb, below=of cam] (det) {YOLO11n + ByteTrack\\lock one person};
\node[capb, below=of det] (crop) {crop $384{\times}512$\\shared frame buffer};
\draw[arr] (cam) -- (det);
\draw[arr] (det) -- (crop);
\node[disp, right=21mm of det]  (flow)  {DIS optical flow\\(CPU, ultrafast)};
\node[disp, right=of flow]  (warp)  {warp latest\\keyframe};
\node[disp, right=of warp]  (fade)  {6-frame cross-fade\\paste through mask};
\node[disp, right=of fade, minimum width=15mm]  (show)  {display\\25--30\,FPS};
\draw[darr] (flow) -- (warp);
\draw[darr] (warp) -- (fade);
\draw[darr] (fade) -- (show);
\node[gen, below=17mm of flow]  (mask) {SCHP + DensePose\\\longmask{}/\uppermask{} mask};
\node[gen, right=of mask]  (enc)  {VAE encode\\person + garment};
\node[gen, right=of enc]   (unet) {CatVTON (Run B)\\LCM 6 steps, no CFG};
\node[gen, right=of unet]  (dec)  {VAE decode\\$\rightarrow$ keyframe};
\draw[garr] (mask) -- (enc);
\draw[garr] (enc) -- (unet);
\draw[garr] (unet) -- (dec);
\node[ms, below=0.8mm of mask] {$\approx$70\,ms};
\node[ms, below=0.8mm of enc]  {$\approx$40\,ms};
\node[ms, below=0.8mm of unet] {$\approx$230\,ms};
\node[ms, below=0.8mm of dec]  {$\approx$45\,ms (full VAE)};
\begin{scope}[on background layer]
\node[lane, fill=orange!5, draw=orange!40, fit=(flow)(show), inner ysep=3.2mm] (dlane) {};
\node[lane, fill=blue!4,  draw=blue!30,   fit=(mask)(dec), inner ysep=5mm] (glane) {};
\end{scope}
\node[lanelabel, text=orange!60!black] at ([xshift=1mm,yshift=-0.8mm]dlane.north west) {Display loop \quad\mdseries every camera frame};
\node[lanelabel, text=blue!50!black]   at ([xshift=1mm,yshift=-0.8mm]glane.north west) {Generator \quad\mdseries asynchronous, one keyframe every 370--490\,ms};
\coordinate (bus) at ([xshift=3.5mm]crop.east);
\draw[darr] (crop.east) -- (bus) |- (flow.west);
\draw[garr] (crop.east) -- (bus) |- (mask.west);
\node[lbl, text=orange!70!black, above=0.5pt of {$(bus |- flow.west)!0.5!(flow.west)$}] {every frame};
\node[lbl, text=blue!55!black,  below=0.5pt of {$(bus |- mask.west)!0.5!(mask.west)$}] {latest frame,\\when idle};
\coordinate (mid) at ($(dlane.south)!0.5!(glane.north)$);
\draw[garr, dashed] (dec.north) -- (dec.north |- mid) -- (warp.south |- mid) -- (warp.south);
\node[lbl, text=blue!50!black, above=0.5pt of {$(dec.north |- mid)!0.5!(warp.south |- mid)$}] {new keyframe enters only through the cross-fade};
\end{tikzpicture}
\caption{\textbf{\ours{} frame path.} The grey column captures, tracks, and crops every camera frame into a shared buffer. The orange lane is the display loop and runs on every frame; the blue lane is the generator and runs asynchronously on the latest frame, producing a try-on keyframe every 370--490\,ms (stage times are for the shipped setting, Table~\ref{tab:speed}). Between keyframes the latest result is warped to the current frame with DIS optical flow, and a new keyframe enters only through a 6-frame cross-fade, so the display never waits for the generator (\akt{}, Section~\ref{sec:akt}).}
\label{fig:pipeline}
\end{figure*}

Figure~\ref{fig:pipeline} shows the frame path. Each $1280\times720$ camera frame goes to YOLO11n~\cite{ultralytics2024yolo11} with ByteTrack~\cite{zhang2022bytetrack}; the user selects one track, and only that person is ever edited, so a second person in frame is left pixel-identical. A $3{:}4$ crop with 15\% margin is resized to $384\times512$. The clothing-agnostic mask is built from SCHP ATR and LIP~\cite{li2020schp} and DensePose~\cite{guler2018densepose} as in CatVTON, with the \longmask{}/\uppermask{} part chosen from the garment (Section~\ref{sec:mask}) and, for \longmask{} garments, cut 40\,px below the hem of the previous keyframe's garment (the hem-aware mask of Section~\ref{sec:mask}, read from the parse the tight matte already computes); we moved the parsers' post-processing to the GPU, which cut the mask stage from 240 to 70\,ms at 99.97\% label agreement. The crop, mask, and garment card go to CatVTON (our fine-tuned Run B attention weights, LCM-LoRA~\cite{luo2023lcmlora} on top, 6 steps, no guidance, full VAE decoder), and the keyframe patch is pasted back through the mask with bicubic resampling. Flow is DIS~\cite{kroeger2016dis} (OpenCV, CPU, ``fast'' preset), chosen because it costs no GPU time; the cross-fade is 6 frames.

\subsection{Measurements}
\label{sec:livemeas}

\begin{table}[t]
\centering\footnotesize
\setlength{\tabcolsep}{3pt}
\caption{\textbf{\AKT{} on a synthetic $1280\times720$ two-person video}, 45\,s headless runs, RTX~4080. ``Age'' is how old the try-on result shown in a frame is; flicker is the mean absolute 8-bit change inside the garment mask between consecutive displayed frames, split into warp-only frames and frames where a new keyframe enters. Base CatVTON \texttt{vitonhd} weights at $384\times512$ unless noted.}
\label{tab:live}
\resizebox{\linewidth}{!}{\begin{tabular}{@{}lrrrrrr@{}}
\toprule
Configuration & FPS & Gen ms & Age ms & Gens & \multicolumn{2}{c}{Flicker} \\
\cmidrule(lr){6-7}
 & & & & /45\,s & warp & key \\
\midrule
LCM 4 + TAESD, DIS flow     & 28.7 & 344 & 545 & 123 & 2.1 & 8.4 \\
\quad same, RAFT-small flow & 21.8 & 354 & 599 & 121 & 2.9 & 8.6 \\
\quad same, no warp         & 29.1 & 329 & 519 & 130 & 0.0 & 9.7 \\
10 steps CFG 2.5 + TAESD, DIS & 14.9 & 599 & 886 & 74 & 3.5 & 12.6 \\
FASHN 1.5, $576{\times}864$, 8 steps, DIS & 4.7 & 1820 & 2515 & 24 & 13.7 & 24.0 \\
\bottomrule
\end{tabular}}
\end{table}

\paragraph{Keyframe warping.}
Table~\ref{tab:live} compares display strategies on a synthetic test video with controlled motion and two people. DIS flow keeps the displayed garment within about 2 grey levels of frame-to-frame change at no GPU cost and leaves the display at 29\,FPS. RAFT-small is not more stable and costs 7\,FPS because it competes with the diffusion model for the GPU. Freezing the keyframe (no warp) has zero warp flicker by construction but the garment visibly detaches from the body, and the keyframe switch becomes larger. The dominant residual artefact is the keyframe switch (8--13 levels), where a freshly sampled result replaces the warped one; the fixed seed and the cross-fade reduce it, and warping the previous \emph{latent} as an SDEdit-style initialisation is the natural next step. Only the locked track is edited: the second person in the video is pixel-identical to the input in every run.

\paragraph{Real camera.}
On a Lenovo FHD webcam at $1280\times720$ with the Run B weights, the setting we ship (LCM 6 steps, full VAE decoder) generates a keyframe in 370--400\,ms including masking, with the display loop at 25\,FPS, and the whole server (tracker, parsers, try-on model, photo-mode model) occupies about 70\% of the card's 16\,GB. We do not include recordings of live sessions in the paper: the people in them are not models, and we did not want to publish their likeness. The quantitative numbers above, Table~\ref{tab:speed}, and the held-out results of Section~\ref{sec:experiments} are the evidence we report instead.

\begin{table*}[t]
\centering\small
\setlength{\tabcolsep}{6pt}
\caption{\textbf{Speed versus print quality} for one $384\times512$ try-on on the RTX~4080 (real camera crops, two people, four garments with text and floral prints; \texttt{vhd} = VITON-HD weights, \texttt{mix} = VITON-HD+DressCode weights at the stated resolution). Verdicts are from zoomed side-by-side inspection.}
\label{tab:speed}
\begin{tabular}{@{}lrrrl@{}}
\toprule
Setting & ms & UNet & Dec. & Print quality \\
\midrule
\texttt{vhd} LCM 4 + TAESD & 305 & 160 & 4 & text blotchy, florals mushy \\
\texttt{vhd} LCM 6 + TAESD & 370 & 230 & 4 & better, text mostly legible \\
\textbf{\texttt{vhd} LCM 6 + full VAE} & \textbf{408} & 226 & 45 & \textbf{legible text, defined florals} \\
\texttt{vhd} 10 DDIM + CFG 2.5 + TAESD & 635 & 496 & 4 & sharp but tinted \\
\texttt{mix} $576{\times}768$ LCM 8 + full VAE & 1023 & 750 & 102 & best, 2.5$\times$ slower \\
\texttt{mix} $576{\times}768$ 8 DDIM + CFG & 1275 & 1049 & 8 & grey tint \\
\texttt{mix} $768{\times}1024$ LCM 6 + TAESD & 1518 & 1201 & 20 & pink tint \\
\bottomrule
\end{tabular}
\end{table*}

\paragraph{Which accelerations keep the garment.}
Table~\ref{tab:speed} summarises the inference study that chose the shipped setting. Three findings transfer beyond our system. (1) At $384\times512$ the tiny TAESD decoder~\cite{taesd}, not the step count, is what destroys printed text and florals; the full VAE decoder costs only 40\,ms more. (2) Two extra consistency steps (4 to 6) recover most of the remaining detail; beyond that, resolution matters more than steps. (3) Classifier-free guidance and the mixed-dataset checkpoint below their native $768\times1024$ add colour tints, so a no-guidance consistency sampler at the training resolution is the right operating point for live use. We also measured \texttt{torch.compile} (reduce-overhead) at only $1.03$--$1.11\times$ for CatVTON after an 80\,s warm-up, and found fp16 and bf16 identical in speed, so neither is worth its cost here. The LCM-LoRA, trained against the original Stable Diffusion attention, still works on top of our fine-tuned attention weights but is slightly softer than on the base.

\begin{table*}[t]
\centering\small
\setlength{\tabcolsep}{6pt}
\caption{\textbf{Latency budget} of one keyframe at $384\times512$ (RTX~4080, LCM 4 steps, TAESD; measured with CUDA synchronisation) and the stage targets that would be needed for a 300\,ms keyframe at 720p.}
\label{tab:budget}
\begin{tabular}{@{}lrrl@{}}
\toprule
Stage & Now ms & Target & How \\
\midrule
capture, track, crop & 20--25 & $\le$15 & TensorRT detector, track every 2nd frame \\
agnostic mask & 70 & $\le$25 & one light parser, run at 512 \\
VAE encode (2 images) & 41 & $\le$15 & tiny encoder; cache garment latent \\
denoiser (4 steps, no CFG) & 138 & $\le$180 & 1--2-step distilled student; cached garment KV \\
decode & 3.5 & $\le$10 & full VAE acceptable at 384, TAESD at 720p \\
paste, flow, display & 10 & $\le$15 & DIS ``ultrafast'' \\
\midrule
total & $\approx$330 (age $\approx$550) & 260--300 & \\
\bottomrule
\end{tabular}
\end{table*}

\paragraph{Where the time goes.}
Table~\ref{tab:budget} breaks a keyframe down by stage. The denoiser is under half of the total; masking and VAE encoding, which are rarely reported, are a third. Reaching a 300\,ms keyframe at 720p would need a 1--2-step distilled try-on student (the consistency LoRA already smears prints at 4 steps), a cached garment key/value path so the garment no longer doubles the attention sequence, and TensorRT preprocessing; \akt{} would remain unchanged. We report these as targets, not results.

\section{Adapting the model to Indian ethnic wear}
\label{sec:method}

\subsection{Base model}
CatVTON~\cite{chong2024catvton} is a Stable Diffusion~1.5 inpainting UNet~\cite{rombach2022ldm} with the text cross-attention removed. The person latent and the garment latent, each $4\times64\times48$ at $384\times512$, are concatenated along the height axis, together with the clothing-agnostic mask (zero on the garment half) and the masked person latent, giving a 9-channel input. Only the self-attention projections (49.6M parameters) are trained; the rest of the UNet is frozen. The loss is the standard $\epsilon$-prediction MSE on the person half of the concatenation. We start from the public \texttt{vitonhd} checkpoint.

\subsection{The \longmask{} mask}
\label{sec:mask}
CatVTON's masker builds the clothing-agnostic region from SCHP parsing (ATR and LIP label sets) and DensePose body parts. Its \uppermask{} part masks torso and arms and keeps everything below the waist; its \overallmask{} part masks the whole body down to the ankles. Neither can express a kurta: the first forbids a hem below the waist, the second erases the trousers that must remain visible.

We add a \longmask{} part: the union of the \uppermask{} region and the DensePose upper-leg parts (front and back thighs), with the lower legs, feet, hands, and face kept. The hidden region thus extends to the knee, where a kurta hem typically lands, and the model is free to draw either garment or trousers inside it. Which mask to use is decided per garment by a keyword rule on the product title or category (\emph{kurta, sherwani, pathani, kurti, anarkali} $\rightarrow$ \longmask{}; \emph{Nehru jacket, waistcoat, short kurta, shirt} $\rightarrow$ \uppermask{}); in the mirror the rule is applied to the garment file name. Using \longmask{} on a T-shirt makes the model invent a skirt, so the rule matters.

\paragraph{Hem-aware two-pass mask.} The \longmask{} mask hides the trousers between the hem and the knee, so the model must guess their colour, and on 16--17\% of held-out pairs it guesses wrong (Section~\ref{sec:experiments}). The hem of the new garment is not known before generation, and estimating it from the card's aspect ratio and the person's shoulder width is too coarse (median error 20\,px, 90th percentile 50\,px at $512$ high on 399 pairs; an underestimate truncates the garment, which is worse than the colour error). We therefore take the hem from a first generation: generate with the full \longmask{} mask, parse the output with SCHP, take the lowest garment row $h$, and generate again with the mask cleared below $h+30$\,px (floored by the hem of the garment the person is actually wearing, so an old long garment is still hidden). The second pass sees the real trousers immediately below the hem and copies them. Offline this doubles the cost (0.8\,s per image); in the mirror it costs nothing, because the previous keyframe is the first pass: its hem is read from the SCHP parse that the tight matte already computes, and the next keyframe's mask is cut at that hem plus 40\,px. Training with the cut mask as augmentation (Run C below) adds nothing measurable, so the fix is inference-only.

\subsection{Self-supervised pairs from on-model photographs}
\label{sec:pairs}
Given an on-model product photograph $I$, we build a triplet (person, garment card, mask):
\begin{enumerate}
\item \textbf{Person.} $I$ letter-boxed to $384\times512$ on white.
\item \textbf{Garment card.} SCHP ATR and LIP are run on $I$; the garment region is the union of the ATR upper-clothes and dress labels and the LIP upper-clothes, dress, coat, and jumpsuit labels. Pixels of the person's hands that overlap the garment leave holes, which are filled with Telea inpainting~\cite{telea2004}. The region is placed on a white $3{:}4$ card. The result is a ``model-worn'' garment image in the person's own pose.
\item \textbf{Mask.} The CatVTON masker is run on $I$ with the part chosen by the keyword rule of Section~\ref{sec:mask}.
\item \textbf{Target.} $I$ itself.
\end{enumerate}
\begin{figure}[t]
\centering
\begin{tikzpicture}[font=\scriptsize, node distance=2.8mm and 3mm,
  box/.style={draw=black!60, line width=0.5pt, rounded corners=1.5pt, align=center, inner sep=2pt, minimum height=6mm, fill=white, text width=23mm},
  src/.style={box, fill=black!4},
  proc/.style={box, fill=blue!8, draw=blue!50!black},
  outb/.style={box, fill=green!10, draw=green!40!black},
  gate/.style={box, fill=red!6, draw=red!50!black, dashed},
  arr/.style={-{Stealth[length=1.5mm]}, line width=0.6pt, black!65}]
\node[src, text width=60mm] (img) {on-model product photograph $I$ (IndoFashion, Atlas)};
\node[gate, below=of img, text width=60mm] (filt) {quality filter: exactly one face blob $\ge$0.2\% of the image, at most two garment blobs $>$15\% of the largest};
\draw[arr] (img) -- (filt);
\coordinate (split) at ([yshift=-2mm]filt.south);
\node[proc, below=5mm of filt, xshift=-27mm] (parse) {SCHP ATR + LIP: upper-clothes, dress, coat, jumpsuit labels};
\node[proc, below=5mm of filt]               (letter) {letter-box to $384{\times}512$ on white};
\node[proc, below=5mm of filt, xshift=27mm]  (masker) {CatVTON masker, part by keyword rule};
\draw[arr] (filt.south) -- (split) -| (parse.north);
\draw[arr] (filt.south) -- (letter.north);
\draw[arr] (filt.south) -- (split) -| (masker.north);
\node[proc, below=of parse]  (inp)  {hand holes $\to$ Telea inpaint; paste on white $3{:}4$ card};
\node[proc, below=of masker] (part) {\longmask{}: kurta, sherwani, anarkali \ldots\\ \uppermask{}: Nehru, shirt \ldots};
\draw[arr] (parse) -- (inp); \draw[arr] (masker) -- (part);
\node[outb, below=of inp]   (ocard)   {\textbf{garment card}\\model-worn, in $I$'s pose};
\node[outb, below=of part]  (omask)   {\textbf{mask}\\\longmask{} or \uppermask{}};
\node[outb] (operson) at (ocard -| letter) {\textbf{person = target}\\$I$ itself};
\draw[arr] (inp) -- (ocard); \draw[arr] (part) -- (omask); \draw[arr] (letter) -- (operson);
\node[below=2mm of operson, font=\tiny, text=black!60, text width=80mm, align=center] {filter removes 4.4\% of IndoFashion and 1.1\% of Atlas candidates $\to$ 23,646 + 1,221 ethnic pairs, mixed with 11,647 VITON-HD flat-lay pairs (Table~\ref{tab:data}); 0.15\,s per pair on the GPU};
\end{tikzpicture}
\caption{\textbf{Self-supervised pair construction.} One on-model photograph yields a (person, card, mask) triplet with the photograph itself as the target, so training is a reconstruction: put the garment back at the right length and draw the trousers the \longmask{} mask hid. The card is cut from the person by the parser and inpainted where the hands occluded it; the mask part comes from the keyword rule. The dashed box is the quality filter.}
\label{fig:pairflow}
\end{figure}
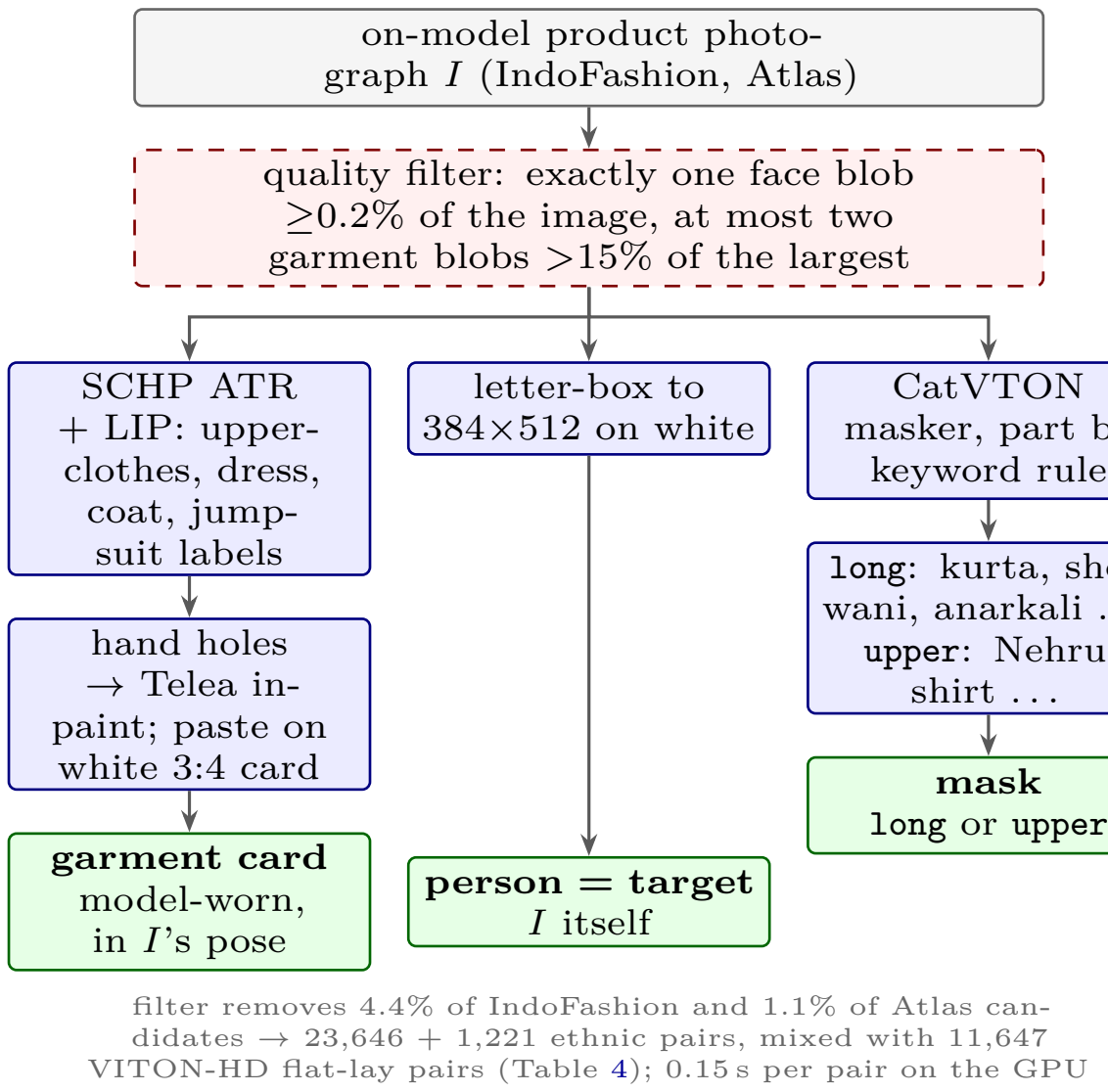

Figure~\ref{fig:pairflow} summarises the construction. Because the garment card is taken from the target, the task is a reconstruction: the model must put the garment back where it was, at the right length, and must draw the trousers that the \longmask{} mask hid. The card is in the person's pose rather than flat, which differs from the inference-time distribution; we mitigate this by always mixing the ethnic pairs with real flat-lay pairs from VITON-HD (Section~\ref{sec:training}) and observe no regression on flat-lay inputs.

\paragraph{Quality filter.}
On-model listings include collages, garment-only images, and group shots. We keep an image only if the parser finds exactly one face blob of at least 0.2\% of the image, and at most two garment blobs larger than 15\% of the largest one. This removed 4.4\% of the IndoFashion and 1.1\% of the Atlas candidates; VITON-HD pairs are kept as is.

\begin{figure}[t]
\centering
\includegraphics[width=\linewidth]{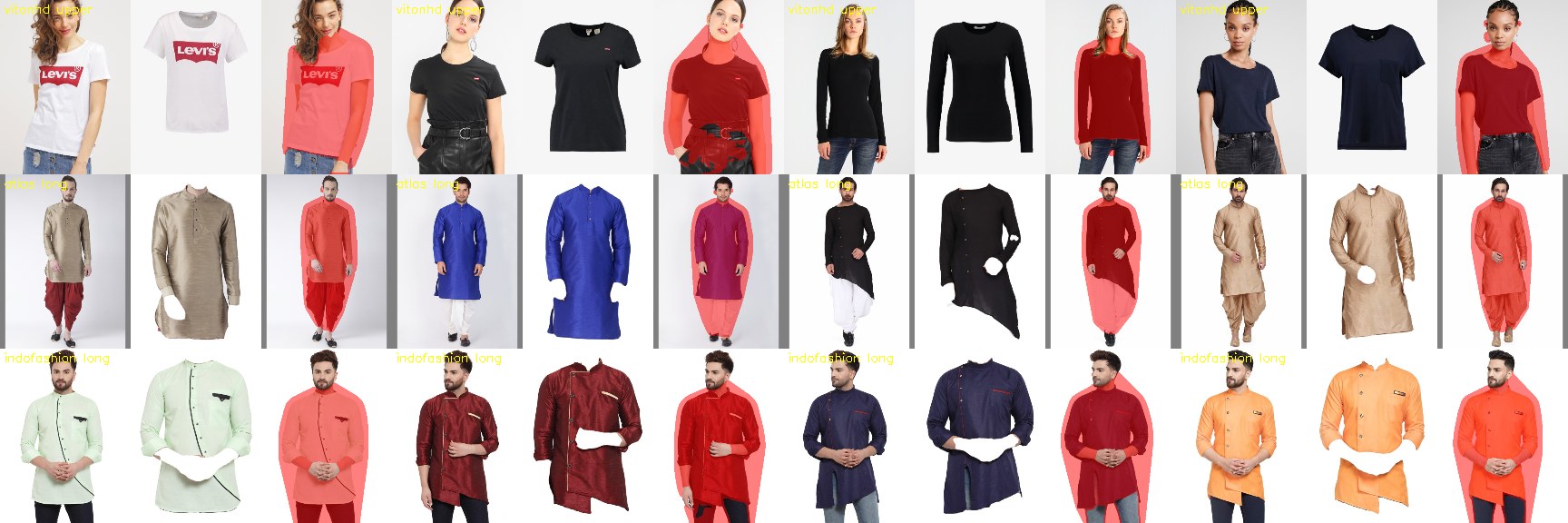}
\caption{\textbf{Training triplets} (person, garment card, mask overlay in red). Top: VITON-HD flat-lay pairs with the \uppermask{} mask. Middle and bottom: Atlas and IndoFashion self-supervised pairs with the \longmask{} mask. The cut-out keeps the garment in the person's pose; hand holes are inpainted.}
\label{fig:pairs}
\end{figure}

\paragraph{Sources.}
Table~\ref{tab:data} lists the three sources. VITON-HD train~\cite{choi2021vitonhd} supplies real flat-lay pairs with the \uppermask{} mask. Atlas~\cite{marqo2024atlas} rows whose titles match ethnic-wear keywords, and the \emph{kurta\_men}, \emph{sherwanis}, \emph{nehru\_jackets}, and \emph{women\_kurta} classes of IndoFashion~\cite{rajput2021indofashion}, supply on-model photographs. IndoFashion ships $\sim$320\,px thumbnails; we re-fetched the full-resolution originals from the listed URLs where still available (about 26k of 32k). Building one pair takes 0.15\,s on the GPU. Figure~\ref{fig:pairs} shows examples.

\begin{table}[t]
\centering\footnotesize
\setlength{\tabcolsep}{2.5pt}
\caption{Training pairs after the quality filter. All sources are public. ``Card'' is the garment conditioning image; ``A / B'' gives the sampling in each run.}
\label{tab:data}
\begin{tabular}{@{}llllrl@{}}
\toprule
Source & Type & Card & Mask & Pairs & A / B \\
\midrule
VITON-HD & Western & flat-lay & \uppermask{} & 11,647 & 1$\times$ / 1$\times$ \\
Atlas & ethnic & cut-out & \longmask{}, \uppermask{} & 1,221 & 501@8$\times$ / 4$\times$ \\
IndoFashion & ethnic & cut-out & \longmask{}, \uppermask{} & 23,646 & -- / 1$\times$ \\
\midrule
Total & & & & 36,514 & \\
\bottomrule
\end{tabular}
\end{table}

\subsection{Training}
\label{sec:training}
We train the attention parameters with AdamW~\cite{loshchilov2019adamw} (learning rate $10^{-5}$, weight decay $10^{-2}$, 200 warm-up steps, cosine decay to 10\%), batch size 8, bf16 autocast with gradient checkpointing, an EMA of the weights with decay 0.999, and the garment card dropped with probability 0.1 so that classifier-free guidance~\cite{ho2022cfg} remains usable. The $\epsilon$-MSE is weighted $2\times$ inside the mask. Ethnic pairs are oversampled to balance the sources (Table~\ref{tab:data}).

Training proceeds in two stages, which we also report as an ablation:
\begin{itemize}
\item \textbf{Run A} (\texttt{ftA}): VITON-HD plus the first 501 Atlas pairs (oversampled $8\times$), 3,000 steps, 68\,min.
\item \textbf{Run B} (\texttt{ftB}): all 36.5k pairs (Atlas oversampled $4\times$), initialised from the EMA weights of Run A, 6,000 steps, 125\,min.
\item \textbf{Run C} (\texttt{ftC}, ablation): Run B's recipe continued for 2,000 steps (42\,min) from Run B's EMA weights with the hem-aware cut applied to 70\% of \longmask{} pairs as augmentation (mask cleared 20--90\,px below the known hem), to test whether the two-pass mask needs matching training. It does not (Table~\ref{tab:main}).
\end{itemize}
Peak memory is 7.1\,GB at 1.25--1.35\,s per step on the RTX~4080, so the whole adaptation (Runs A and B) fits in 3.2\,GPU-hours on the same card that runs the mirror. The final training loss was 0.063--0.065 for Run B versus 0.070 for Run A.

\section{Adaptation experiments}
\label{sec:experiments}

\begin{table*}[!t]
\centering\footnotesize
\caption{\textbf{Held-out reconstruction} (200 pairs per source, 10 DDIM steps, CFG 2.5, $384\times512$; mean, with the per-pair standard deviation in small type for LPIPS and LBF). Lower LPIPS and higher PSNR/SSIM are better. Lower-band fidelity (LBF) is defined for \longmask{} pairs only and is always measured on the \longmask{} band; ``fail'' is the fraction of \longmask{} pairs with LBF $<0.3$, i.e.\ the trousers were repainted in another colour. ``+hem'' is the two-pass hem-aware mask of Section~\ref{sec:mask}, which changes nothing on the \uppermask{}-only VITON-HD pairs. Run B improves every ethnic-wear metric over the base, Run A, and FASHN without regressing on VITON-HD; the hem-aware mask then removes most of the remaining trouser failures, and training with the cut mask (Run C) adds nothing. $^\dagger$FASHN uses its own parser-based mask for the \texttt{tops} category; garments its parser labels as outerwear (some sherwanis, Nehru jackets) are left unmasked and copied, which flatters its scores on those pairs.}
\label{tab:main}
\setlength{\tabcolsep}{2.6pt}
\newcommand{\sd}[1]{{\scriptsize$\pm$#1}}
\resizebox{\textwidth}{!}{%
\begin{tabular}{@{}l cccc c cccc c ccc@{}}
\toprule
& \multicolumn{5}{c}{Atlas (ethnic)} & \multicolumn{5}{c}{IndoFashion (ethnic)} & \multicolumn{3}{c}{VITON-HD (Western)} \\
\cmidrule(lr){2-6}\cmidrule(lr){7-11}\cmidrule(lr){12-14}
Checkpoint & PSNR$\uparrow$ & SSIM$\uparrow$ & LPIPS$\downarrow$ & LBF$\uparrow$ & fail$\downarrow$ & PSNR$\uparrow$ & SSIM$\uparrow$ & LPIPS$\downarrow$ & LBF$\uparrow$ & fail$\downarrow$ & PSNR$\uparrow$ & SSIM$\uparrow$ & LPIPS$\downarrow$ \\
\midrule
FASHN 1.5 (masked)$^\dagger$ & 19.49 & 0.841 & 0.152\sd{.08} & 0.374\sd{.37} & 54\% & 20.82 & 0.840 & 0.119\sd{.06} & 0.478\sd{.36} & 40\% & 20.59 & \textbf{0.855} & \textbf{0.116}\sd{.04} \\
CatVTON base (\texttt{vitonhd}) & 20.20 & 0.835 & 0.135\sd{.07} & 0.389\sd{.21} & 33\% & 20.25 & 0.821 & 0.137\sd{.07} & 0.417\sd{.24} & 32\% & 20.30 & 0.844 & 0.130\sd{.06} \\
Run A (\texttt{ftA}, EMA)        & 21.65 & 0.850 & 0.083\sd{.04} & 0.503\sd{.24} & 19\% & 21.57 & 0.829 & 0.088\sd{.05} & 0.497\sd{.25} & 21\% & 20.62 & 0.844 & 0.114\sd{.06} \\
Run B (\texttt{ftB}, EMA)        & 22.65 & 0.862 & \textbf{0.073}\sd{.04} & 0.547\sd{.25} & 17\% & 22.60 & 0.845 & 0.077\sd{.05} & 0.558\sd{.26} & 16\% & 20.32 & 0.842 & 0.117\sd{.06} \\
\midrule
Run B + hem (two-pass)           & 23.84 & 0.869 & \textbf{0.065}\sd{.04} & 0.713\sd{.20} & 4.2\% & 23.48 & 0.851 & 0.070\sd{.05} & 0.715\sd{.20} & \textbf{2.5\%} & 20.31 & 0.842 & 0.117\sd{.06} \\
Run C (\texttt{ftC}, EMA) + hem  & \textbf{23.97} & \textbf{0.871} & \textbf{0.065}\sd{.04} & \textbf{0.716}\sd{.20} & \textbf{3.2\%} & \textbf{23.66} & \textbf{0.853} & \textbf{0.069}\sd{.05} & \textbf{0.720}\sd{.20} & 3.1\% & 20.15 & 0.840 & 0.120\sd{.06} \\
\bottomrule
\end{tabular}}
\end{table*}

\begin{figure*}[!t]
\centering
\includegraphics[width=0.7\textwidth]{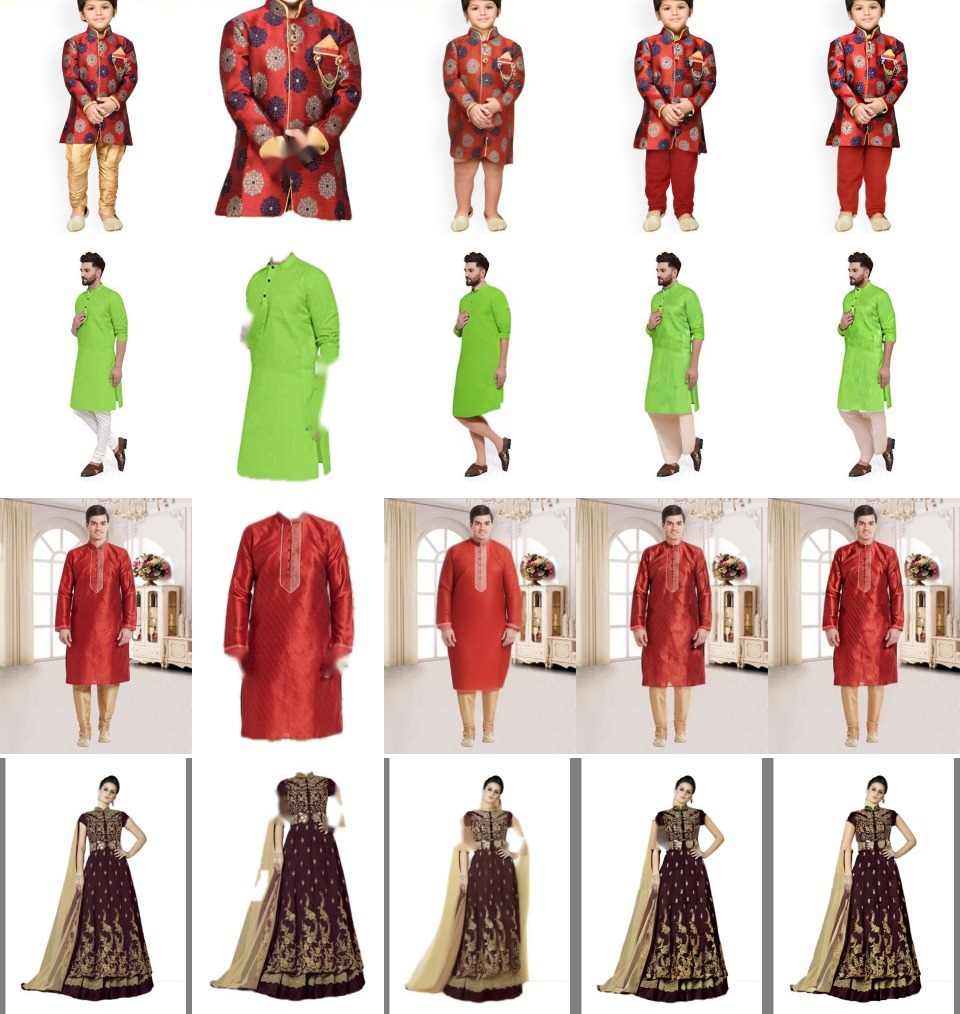}
\caption{\textbf{Held-out ethnic-wear reconstructions.} Columns: person, garment card, CatVTON base, Run A, Run B. Row 1: the base model renders the boy's sherwani as a dress with bare legs; Runs A and B restore the trousers (the invented colour, red instead of gold, is the trouser-recolouring failure that the hem-aware mask of Section~\ref{sec:mask} removes; Figure~\ref{fig:hem}). Row 2: a kurta over white trousers; the base invents bare legs. Row 3: the base model loses the body shape and hem; Run B recovers both. Row 4: an anarkali with \emph{dupatta}; the base makes the dupatta a second skirt, Run B keeps it as a drape.}
\label{fig:hero}
\end{figure*}

\subsection{Setup}
\textbf{Held-out set.} We hold out 200 pairs per source from a seeded shuffle of the index (not used in training; the first 24 of each source were used during development, and the numbers on that subset are consistent with the full set). Every held-out pair is self-paired, so the original photograph is the ground truth for the reconstruction.

\textbf{Metrics.} PSNR, SSIM~\cite{wang2004ssim}, and LPIPS (AlexNet)~\cite{zhang2018lpips} between the output and the photograph. For \longmask{} pairs we add a \emph{lower-band fidelity} score, $1 - \mathrm{MAE}/60$ over the lower 30\% of the mask bounding box, which measures whether the hem and trouser region was reconstructed rather than invented; it is clipped to $[0,1]$, and we call a pair a \emph{trouser failure} when it is below 0.3, which in practice means the trousers were repainted in a different colour. The band is always the \longmask{} band, whichever mask was used to generate.

\textbf{Sampling.} 10 DDIM~\cite{song2021ddim} steps, classifier-free guidance 2.5, $384\times512$, fixed seed, full VAE decoder, for all checkpoints.

\textbf{Baselines.} (i) The base CatVTON \texttt{vitonhd} checkpoint with the same \longmask{} mask, which isolates the effect of training. (ii) FASHN VTON 1.5~\cite{fashn2025vton}, a pixel-space MMDiT try-on model that is the strongest open model we could run on 16\,GB, at 20 steps and guidance 1.5 at its native $576\times864$, with the garment card passed as a flat-lay and the output composited back through our mask so the untouched region is identical. FASHN has no external-mask input, so it hides the worn garment with its own parser (category \texttt{tops}). Its default \emph{maskless} mode must not be used for a self-paired evaluation: it sees the worn garment and reproduces the photograph almost exactly (LPIPS 0.033 on IndoFashion), which measures copying, not try-on. IDM-VTON~\cite{choi2024idmvton} was not run because its 29\,GB of weights exceed our download budget.

\subsection{Quantitative results}
Table~\ref{tab:main} gives the main result on 200 held-out pairs per source. On the two ethnic-wear sources LPIPS falls by 46\% (Atlas) and 44\% (IndoFashion) from base to Run B, PSNR rises by about 2.4\,dB, and lower-band fidelity, the metric most sensitive to the trousers-under-hem problem, rises from 0.39/0.42 to 0.55/0.56, with the trouser-failure rate halved from 33\%/32\% to 17\%/16\%. The per-pair standard deviations are large (0.04--0.07 LPIPS), but with $n=200$ the standard error of each mean is below 0.005 LPIPS, so the gaps between rows are many standard errors wide. On VITON-HD, which contains only Western tops with flat-lay cards, LPIPS improves slightly (0.130 to 0.117) and PSNR/SSIM are flat, so the adaptation does not cost the original capability.

\begin{figure}[!t]
\centering
\includegraphics[width=\linewidth]{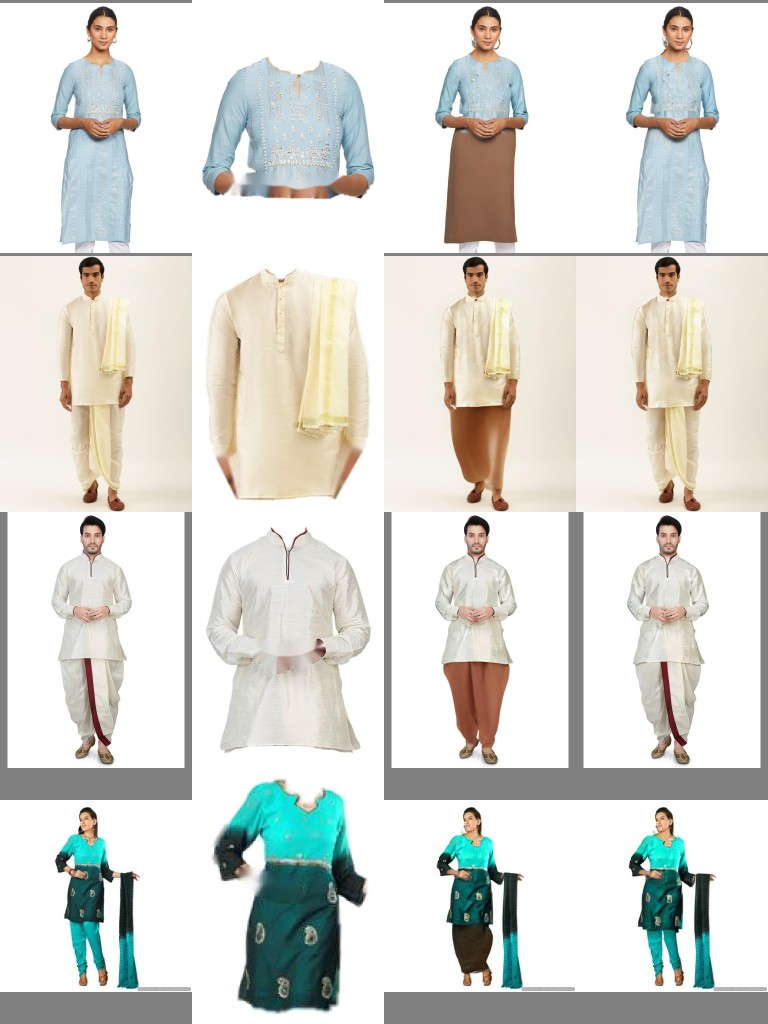}
\caption{\textbf{Hem-aware two-pass mask} (Run B, held-out pairs). Columns: person, card, output with the plain \longmask{} mask, output with the mask cut 30\,px below the hem found in the first pass. The plain mask hides the trousers down to the knee and the model repaints them in an invented colour (brown or ochre under cream and white kurtas, teal under the green suit); the second pass sees the real trousers just below the hem and keeps them.}
\label{fig:hem}
\end{figure}

\paragraph{Hem-aware mask.} The two-pass mask (Section~\ref{sec:mask}) applied to Run B at inference removes most of what is left of the trouser problem: the failure rate drops from 17\% to 4.2\% on Atlas and from 16\% to 2.5\% on IndoFashion, lower-band fidelity rises to 0.71 on both, and LPIPS and PSNR improve further (0.065/0.070, $+1.2$/$+0.9$\,dB) because the band below the hem is now copied rather than repainted. Run C, which saw the cut mask during training, is within noise of Run B under the same two-pass inference (differences of 0.001 LPIPS and about 1\% in failure rate, in both directions) and is slightly worse on VITON-HD, so we ship Run B and treat the hem-aware mask as a pure inference-time fix. Figure~\ref{fig:hem} shows the effect. The residual 3--4\% are pairs where the first pass itself is wrong: the model extends the garment into a floor-length dress, and the parsed hem then lands below the knee, which leaves the mask unchanged.

\paragraph{FASHN as a second baseline.} FASHN is the stronger model on Western tops (LPIPS 0.116 versus 0.130 for base CatVTON, and the best SSIM on VITON-HD) but shows the same blind spot on ethnic wear in a different form (appendix Figure~\ref{fig:fashn}): its \texttt{tops} prior truncates a knee-length kurta or anarkali to a hip-length top and invents trousers or leggings below it, so hem length is lost rather than the trousers. Its lower-band fidelity on Atlas (0.37) and its trouser-failure rates (54\% and 40\%) are the worst of all rows, and its per-pair spread is the widest (LBF std 0.37), because it either copies a pair its parser did not mask or destroys the hem. Run B beats it on every aggregate ethnic-wear metric (LPIPS 0.073 versus 0.152 on Atlas, 0.077 versus 0.119 on IndoFashion) at about $20\times$ lower inference cost (0.4\,s versus 8.5\,s per image on the same GPU), matches it on VITON-HD LPIPS (0.117 versus 0.116), and the ethnic-wear gap is understated because the pairs FASHN's parser leaves unmasked are effectively free points for it.

\begin{table*}[!t]
\centering
\footnotesize
\setlength{\tabcolsep}{5pt}
\caption{\textbf{Per-garment-type breakdown on the 200 IndoFashion held-out pairs.} LPIPS (lower is better), lower-band fidelity (LBF, higher is better) and trouser-failure rate (LBF $<0.3$) for FASHN, base CatVTON, Run B, and Run B with the hem-aware mask. Nehru jackets use the \uppermask{} mask, so LBF is undefined and the hem-aware mask changes nothing. $^\dagger$FASHN's parser leaves some sherwanis and Nehru jackets unmasked, so part of its score on those rows is copying.}
\label{tab:pertype}
\begin{tabular}{lr cccc cccc cccc}
\toprule
& & \multicolumn{4}{c}{LPIPS $\downarrow$} & \multicolumn{4}{c}{LBF $\uparrow$} & \multicolumn{4}{c}{fail $\downarrow$} \\
\cmidrule(lr){3-6}\cmidrule(lr){7-10}\cmidrule(lr){11-14}
Type & $n$ & FASHN & Base & Run B & B+hem & FASHN & Base & Run B & B+hem & FASHN & Base & Run B & B+hem \\
\midrule
Women's kurta & 94 & 0.138 & 0.128 & 0.063 & \textbf{0.061} & 0.291 & 0.463 & 0.627 & \textbf{0.660} & 61\% & 20\% & 6\% & \textbf{3\%} \\
Men's kurta   & 51 & 0.098 & 0.139 & 0.082 & \textbf{0.066} & 0.727 & 0.381 & 0.503 & \textbf{0.785} & 12\% & 43\% & 26\% & \textbf{2\%} \\
Sherwani$^\dagger$ & 14 & 0.105 & 0.130 & 0.088 & \textbf{0.060} & 0.825 & 0.240 & 0.292 & \textbf{0.836} & 7\% & 71\% & 43\% & \textbf{0\%} \\
Nehru jacket$^\dagger$ & 41 & 0.105 & 0.156 & \textbf{0.099} & \textbf{0.099} & -- & -- & -- & -- & -- & -- & -- & -- \\
\bottomrule
\end{tabular}
\end{table*}

\paragraph{Where the lead is.} Table~\ref{tab:pertype} splits the 200 IndoFashion pairs by garment type. The gain of fine-tuning alone is concentrated in women's kurtas and anarkalis, 94 of the 200 pairs, where Run B reaches LPIPS 0.063 against FASHN's 0.138 and keeps the hem (LBF 0.63 versus 0.29; FASHN fails the trouser band on 61\% of them by cutting the garment short); this is the knee-length-over-trousers silhouette the \longmask{} mask and the training pairs were built for. The breakdown also shows where Run B's own trouser failure lives: under the plain \longmask{} mask it hits 26\% of men's kurtas and 43\% of sherwanis against 6\% of women's kurtas, which is why FASHN, whose parser leaves part of the men's wear unmasked and copied, scores higher LBF on those two rows. The hem-aware mask brings men's kurtas to 2\% and sherwanis to 0\% (LBF 0.79 and 0.84), after which Run B is ahead of FASHN in LPIPS on every type, including the Nehru jackets that no mask change touches (0.099 versus 0.105).

\paragraph{Effect of the second stage.} Run A, trained on only 501 ethnic pairs oversampled, already captures most of the gain on Atlas (LPIPS 0.135 to 0.083) and, notably, generalises to IndoFashion, which it never saw (0.137 to 0.088). Run B's 23.6k IndoFashion pairs add a further 0.010--0.011 LPIPS and 0.04--0.06 lower-band fidelity on both sources. The marginal return of the second stage is therefore small in LPIPS but visible in the hem region, which is where the failure lives.

\subsection{Qualitative results}
Figure~\ref{fig:hero} shows held-out examples. The failure mode of the base model is consistent: with the \longmask{} mask it treats the hidden region as a dress and fills the lower part with skin, the sherwani is shortened, and the anarkali's dupatta becomes an extra skirt layer. After Run A the kurta hem lands at the right length and trousers appear below it; after Run B the trousers are drawn with cleaner boundaries, sherwani brocade and embroidery are preserved, Nehru jackets keep the shirt sleeves of the kurta underneath (Figure~\ref{fig:grid_indo}), and anarkali flare and dupattas survive. On VITON-HD inputs (appendix Figure~\ref{fig:grid_vhd}) the three checkpoints are visually indistinguishable, as the numbers suggest. In the mirror, the Run B weights replace the base weights with no change to the \akt{} settings; the shipped keyframe time of 370--400\,ms in Section~\ref{sec:livemeas} is measured with them.

\section{Limitations}
\label{sec:limitations}
\textbf{Live numbers.} The flicker and warp comparison in Table~\ref{tab:live} was measured on a synthetic video with small rotations; side views, crossed arms, and fast turns are untested, and only the generation time, frame rate, and memory were measured on the real camera. The keyframe switch remains visible at 8--13 grey levels. \textbf{Hidden trouser colour.} The \longmask{} mask hides the band of trousers between waist and knee, so the model must guess their colour; it usually copies the visible lower legs but on 16--17\% of pairs picks a plausible alternative (red instead of gold in Figure~\ref{fig:hero}, row 1). The two-pass hem-aware mask reduces this to 3--4\% of pairs; the rest are first-pass hem errors, and a direct hem estimate from the card was not accurate enough to replace the first pass. In the mirror the hem is taken from the previous keyframe, so the first keyframe after a garment change still uses the full mask. \textbf{Self-paired evaluation.} Our held-out pairs reconstruct the garment the person is already wearing, which is easier than transferring a different garment and does not measure how faithfully a new garment's print or cut is reproduced; a cross-pair evaluation and a human preference study are needed, and, as the FASHN maskless result shows, any baseline must hide the worn garment or the metric rewards copying. \textbf{Sample size.} 200 held-out pairs per source with per-pair standard deviations are reported; the sherwani and Nehru-jacket classes in the per-type breakdown remain small (14 and 41 pairs). \textbf{Card domain gap.} Training cards are in-pose cut-outs while deployment cards are often flat-lays; the VITON-HD mixing appears to bridge this, but we have not measured it. \textbf{Scope.} We cover men's kurta, sherwani, Nehru jackets and women's kurta/anarkali; sarees, lehengas and dhotis, which need different masks, are untouched. \textbf{Data licensing.} IndoFashion images are redistributed under its terms and the re-fetched originals are retailer imagery; any release would consist of the construction code and indices rather than the images.

\section{Conclusion}
A live try-on mirror does not need a faster diffusion model; it needs the display and the generator to stop sharing a clock. \AKT{} gives a camera-rate mirror from a model that takes 400\,ms per result, on one 16\,GB GPU, and we have measured what the remaining artefacts are and where the remaining time goes. The same GPU was enough to teach the model a garment silhouette its training data had never shown it: a thigh-covering mask, parser cut-outs from on-model listings, and 3.2 GPU-hours of attention-only fine-tuning turn a model that paints bare legs under a kurta into one that draws the trousers, and a second pass with the mask cut at the hem it just drew keeps their colour. We hope both recipes, the display scheme and the data recipe, are reused for other garments and other markets, and that the evaluation gaps we list are closed by larger held-out sets, real-camera flicker studies, and human preference tests.

\paragraph{Acknowledgements.} We thank the School of Electronics Engineering (SOEE), KIIT Deemed to be University, for providing the compute and GPU resources used to carry out this research.

{\small
\bibliographystyle{plainnat}
\bibliography{refs}

\begin{thebibliography}{31}
\providecommand{\natexlab}[1]{#1}
\providecommand{\url}[1]{\texttt{#1}}
\expandafter\ifx\csname urlstyle\endcsname\relax
  \providecommand{\doi}[1]{doi: #1}\else
  \providecommand{\doi}{doi: \begingroup \urlstyle{rm}\Url}\fi

\bibitem[Bohan(2023)]{taesd}
Ollin Bohan.
\newblock {TAESD}: Tiny {AutoEncoder} for {Stable Diffusion}.
\newblock \url{https://github.com/madebyollin/taesd}, 2023.

\bibitem[Choi et~al.(2021)Choi, Park, Lee, and Choo]{choi2021vitonhd}
Seunghwan Choi, Sunghyun Park, Minsoo Lee, and Jaegul Choo.
\newblock {VITON-HD}: High-resolution virtual try-on via misalignment-aware
  normalization.
\newblock In \emph{CVPR}, 2021.

\bibitem[Choi et~al.(2024)Choi, Kwak, Lee, Choi, and Shin]{choi2024idmvton}
Yisol Choi, Sangkyung Kwak, Kyungmin Lee, Hyungwon Choi, and Jinwoo Shin.
\newblock Improving diffusion models for authentic virtual try-on in the wild.
\newblock In \emph{ECCV}, 2024.

\bibitem[Chong et~al.(2024)Chong, Dong, Li, Zhang, Zhang, Zhang, Zhao, and
  Liang]{chong2024catvton}
Zheng Chong, Xiao Dong, Haoxiang Li, Shiyue Zhang, Wenqing Zhang, Xujie Zhang,
  Hanqing Zhao, and Xiaodan Liang.
\newblock {CatVTON}: Concatenation is all you need for virtual try-on with
  diffusion models.
\newblock \emph{arXiv preprint arXiv:2407.15886}, 2024.

\bibitem[Chong et~al.(2025)Chong, Zhang, Zhang, Zheng, Dong, Li, Wei, Xu, Lin,
  Ye, and Liang]{chong2025catv2ton}
Zheng Chong, Wenqing Zhang, Shiyue Zhang, Jun Zheng, Xiao Dong, Haoxiang Li,
  Yiling Wei, Chengcheng Xu, Yuxin Lin, Fei Ye, and Xiaodan Liang.
\newblock {CatV2TON}: Taming diffusion transformers for vision-based virtual
  try-on with temporal concatenation.
\newblock \emph{arXiv preprint arXiv:2501.11325}, 2025.

\bibitem[Fang et~al.(2024)Fang, Zhai, Su, Song, Zhu, Wang, Chen, Liu, Cao, and
  Zha]{fang2024vivid}
Zixun Fang, Wei Zhai, Aimin Su, Hongliang Song, Kai Zhu, Mao Wang, Yu~Chen,
  Zhiheng Liu, Yang Cao, and Zheng-Jun Zha.
\newblock {ViViD}: Video virtual try-on using diffusion models.
\newblock \emph{arXiv preprint arXiv:2405.11794}, 2024.

\bibitem[{FASHN AI}(2025)]{fashn2025vton}
{FASHN AI}.
\newblock {FASHN VTON 1.5}: Open-weights virtual try-on model.
\newblock \url{https://github.com/fashn-AI/fashn-vton-1.5}, 2025.
\newblock Accessed 2025-11.

\bibitem[G{\"u}ler et~al.(2018)G{\"u}ler, Neverova, and
  Kokkinos]{guler2018densepose}
R{\i}za~Alp G{\"u}ler, Natalia Neverova, and Iasonas Kokkinos.
\newblock {DensePose}: Dense human pose estimation in the wild.
\newblock In \emph{CVPR}, 2018.

\bibitem[Han et~al.(2018)Han, Wu, Wu, Yu, and Davis]{han2018viton}
Xintong Han, Zuxuan Wu, Zhe Wu, Ruichi Yu, and Larry~S Davis.
\newblock {VITON}: An image-based virtual try-on network.
\newblock In \emph{CVPR}, 2018.

\bibitem[Ho and Salimans(2022)]{ho2022cfg}
Jonathan Ho and Tim Salimans.
\newblock Classifier-free diffusion guidance.
\newblock \emph{arXiv preprint arXiv:2207.12598}, 2022.

\bibitem[Huang et~al.(2025)Huang, Li, He, Zhou, and
  Shechtman]{huang2025selfforcing}
Xun Huang, Zhengqi Li, Guande He, Mingyuan Zhou, and Eli Shechtman.
\newblock Self forcing: Bridging the train-test gap in autoregressive video
  diffusion.
\newblock \emph{arXiv preprint arXiv:2506.08009}, 2025.

\bibitem[Jocher and Qiu(2024)]{ultralytics2024yolo11}
Glenn Jocher and Jing Qiu.
\newblock Ultralytics {YOLO11}.
\newblock \url{https://github.com/ultralytics/ultralytics}, 2024.

\bibitem[Kim et~al.(2024)Kim, Gu, Park, Park, and Choo]{kim2024stableviton}
Jeongho Kim, Guojung Gu, Minho Park, Sunghyun Park, and Jaegul Choo.
\newblock {StableVITON}: Learning semantic correspondence with latent diffusion
  model for virtual try-on.
\newblock In \emph{CVPR}, 2024.

\bibitem[Kodaira et~al.(2023)Kodaira, Xu, Hazama, Yoshimoto, Ohno, Mitsuhori,
  Sugano, Cho, Liu, and Keutzer]{kodaira2023streamdiffusion}
Akio Kodaira, Chenfeng Xu, Toshiki Hazama, Takanori Yoshimoto, Kohei Ohno,
  Shogo Mitsuhori, Soichi Sugano, Hanying Cho, Zhijian Liu, and Kurt Keutzer.
\newblock {StreamDiffusion}: A pipeline-level solution for real-time
  interactive generation.
\newblock \emph{arXiv preprint arXiv:2312.12491}, 2023.

\bibitem[Kroeger et~al.(2016)Kroeger, Timofte, Dai, and
  Van~Gool]{kroeger2016dis}
Till Kroeger, Radu Timofte, Dengxin Dai, and Luc Van~Gool.
\newblock Fast optical flow using dense inverse search.
\newblock In \emph{ECCV}, 2016.

\bibitem[Li et~al.(2020)Li, Xu, Wei, and Yang]{li2020schp}
Peike Li, Yunqiu Xu, Yunchao Wei, and Yi~Yang.
\newblock Self-correction for human parsing.
\newblock \emph{IEEE Transactions on Pattern Analysis and Machine
  Intelligence}, 44\penalty0 (6):\penalty0 3260--3271, 2020.

\bibitem[Loshchilov and Hutter(2019)]{loshchilov2019adamw}
Ilya Loshchilov and Frank Hutter.
\newblock Decoupled weight decay regularization.
\newblock In \emph{ICLR}, 2019.

\bibitem[Luo et~al.(2023{\natexlab{a}})Luo, Tan, Huang, Li, and
  Zhao]{luo2023lcm}
Simian Luo, Yiqin Tan, Longbo Huang, Jian Li, and Hang Zhao.
\newblock Latent consistency models: Synthesizing high-resolution images with
  few-step inference.
\newblock \emph{arXiv preprint arXiv:2310.04378}, 2023{\natexlab{a}}.

\bibitem[Luo et~al.(2023{\natexlab{b}})Luo, Tan, Patil, Gu, von Platen, Passos,
  Huang, Li, and Zhao]{luo2023lcmlora}
Simian Luo, Yiqin Tan, Suraj Patil, Daniel Gu, Patrick von Platen,
  Apolin{\'a}rio Passos, Longbo Huang, Jian Li, and Hang Zhao.
\newblock {LCM-LoRA}: A universal {Stable-Diffusion} acceleration module.
\newblock \emph{arXiv preprint arXiv:2311.05556}, 2023{\natexlab{b}}.

\bibitem[{Marqo}(2024)]{marqo2024atlas}
{Marqo}.
\newblock {Marqo Atlas}: A multilingual e-commerce image--text dataset.
\newblock \url{https://huggingface.co/datasets/Marqo/atlas}, 2024.
\newblock Accessed 2025-11.

\bibitem[Morelli et~al.(2022)Morelli, Fincato, Cornia, Landi, Cesari, and
  Cucchiara]{morelli2022dresscode}
Davide Morelli, Matteo Fincato, Marcella Cornia, Federico Landi, Fabio Cesari,
  and Rita Cucchiara.
\newblock {Dress Code}: High-resolution multi-category virtual try-on.
\newblock In \emph{CVPR Workshops}, 2022.

\bibitem[Rajput and Aneja(2021)]{rajput2021indofashion}
Pranjal~Singh Rajput and Shivangi Aneja.
\newblock {IndoFashion}: Apparel classification for {Indian} ethnic clothes.
\newblock In \emph{CVPR Workshops}, 2021.

\bibitem[Rombach et~al.(2022)Rombach, Blattmann, Lorenz, Esser, and
  Ommer]{rombach2022ldm}
Robin Rombach, Andreas Blattmann, Dominik Lorenz, Patrick Esser, and Bj{\"o}rn
  Ommer.
\newblock High-resolution image synthesis with latent diffusion models.
\newblock In \emph{CVPR}, 2022.

\bibitem[Song et~al.(2021)Song, Meng, and Ermon]{song2021ddim}
Jiaming Song, Chenlin Meng, and Stefano Ermon.
\newblock Denoising diffusion implicit models.
\newblock In \emph{ICLR}, 2021.

\bibitem[Telea(2004)]{telea2004}
Alexandru Telea.
\newblock An image inpainting technique based on the fast marching method.
\newblock \emph{Journal of Graphics Tools}, 9\penalty0 (1):\penalty0 23--34,
  2004.

\bibitem[Wang et~al.(2018)Wang, Zheng, Liang, Chen, Lin, and
  Yang]{wang2018cpvton}
Bochao Wang, Huabin Zheng, Xiaodan Liang, Yimin Chen, Liang Lin, and Meng Yang.
\newblock Toward characteristic-preserving image-based virtual try-on network.
\newblock In \emph{ECCV}, 2018.

\bibitem[Wang et~al.(2004)Wang, Bovik, Sheikh, and Simoncelli]{wang2004ssim}
Zhou Wang, Alan~C Bovik, Hamid~R Sheikh, and Eero~P Simoncelli.
\newblock Image quality assessment: from error visibility to structural
  similarity.
\newblock \emph{IEEE Transactions on Image Processing}, 13\penalty0
  (4):\penalty0 600--612, 2004.

\bibitem[Xu et~al.(2025)Xu, Gu, Chen, and Chen]{xu2024ootdiffusion}
Yuhao Xu, Tao Gu, Weifeng Chen, and Chengcai Chen.
\newblock {OOTDiffusion}: Outfitting fusion based latent diffusion for
  controllable virtual try-on.
\newblock In \emph{AAAI}, 2025.

\bibitem[Zhang et~al.(2018)Zhang, Isola, Efros, Shechtman, and
  Wang]{zhang2018lpips}
Richard Zhang, Phillip Isola, Alexei~A Efros, Eli Shechtman, and Oliver Wang.
\newblock The unreasonable effectiveness of deep features as a perceptual
  metric.
\newblock In \emph{CVPR}, 2018.

\bibitem[Zhang et~al.(2022)Zhang, Sun, Jiang, Yu, Weng, Yuan, Luo, Liu, and
  Wang]{zhang2022bytetrack}
Yifu Zhang, Peize Sun, Yi~Jiang, Dongdong Yu, Fucheng Weng, Zehuan Yuan, Ping
  Luo, Wenyu Liu, and Xinggang Wang.
\newblock {ByteTrack}: Multi-object tracking by associating every detection
  box.
\newblock In \emph{ECCV}, 2022.

\bibitem[Zhu et~al.(2023)Zhu, Yang, Zhu, Reda, Chan, Saharia, Norouzi, and
  Kemelmacher-Shlizerman]{zhu2023tryondiffusion}
Luyang Zhu, Dawei Yang, Tyler Zhu, Fitsum Reda, William Chan, Chitwan Saharia,
  Mohammad Norouzi, and Ira Kemelmacher-Shlizerman.
\newblock {TryOnDiffusion}: A tale of two {UNets}.
\newblock In \emph{CVPR}, 2023.

\end{thebibliography}
}

\appendix
\raggedbottom
\section{Reproducibility details}
\begin{itemize}
\item Hardware: NVIDIA RTX~4080 16\,GB, Intel i9-14900K, 62\,GB RAM; PyTorch 2.5.1, CUDA 12.1, diffusers 0.32.2; Lenovo FHD webcam, $1280\times720$ MJPG at 30\,FPS.
\item Mirror: YOLO11n at 640 with ByteTrack; crop aspect 3:4, margin 15\%, width 384; hem-aware mask margin 40\,px; DIS flow ``fast'' preset on CPU; cross-fade 6 frames; hold mode 4\,s with motion trigger at 25\% of crop width; fixed seed per track.
\item Base weights: CatVTON \texttt{vitonhd} attention checkpoint (public). Human parser: SCHP ATR and LIP; DensePose from the CatVTON masker.
\item Pair builder: $384\times512$ triplets, 0.15\,s per pair; quality filter as in Section~\ref{sec:pairs}.
\item Training: AdamW lr $10^{-5}$, wd $10^{-2}$, warm-up 200, cosine to 10\%, batch 8, bf16, gradient checkpointing, EMA 0.999, card dropout 0.1, mask weight 2.0, seed 0. Run A 3,000 steps (68\,min); Run B 6,000 steps from Run A EMA (125\,min); Run C 2,000 steps from Run B EMA with hem cut on 70\% of \longmask{} pairs, jitter 20--90\,px (42\,min); peak 7.1\,GB.
\item Evaluation: 200 held-out pairs per source from a seeded shuffle, 10 DDIM steps, CFG 2.5, full VAE, LPIPS-Alex; two-pass hem mask: SCHP on the first output, mask cleared below hem $+30$\,px (\texttt{train/eval\_full.py}); mirror: hem of the previous keyframe $+40$\,px (\texttt{hem\_margin} in the config). FASHN: 20 steps, guidance 1.5, \texttt{tops}, segmentation on, flat-lay card, output composited through our mask, 8.5\,s per image. Per-type breakdown from the same 200 IndoFashion pairs.
\end{itemize}

\section{Additional qualitative results}
Figures~\ref{fig:grid_indo}--\ref{fig:fashn} show the full IndoFashion and VITON-HD evaluation grids for the three CatVTON checkpoints, and FASHN VTON 1.5 on the same ethnic-wear pairs.
\begin{center}
\includegraphics[width=0.72\linewidth]{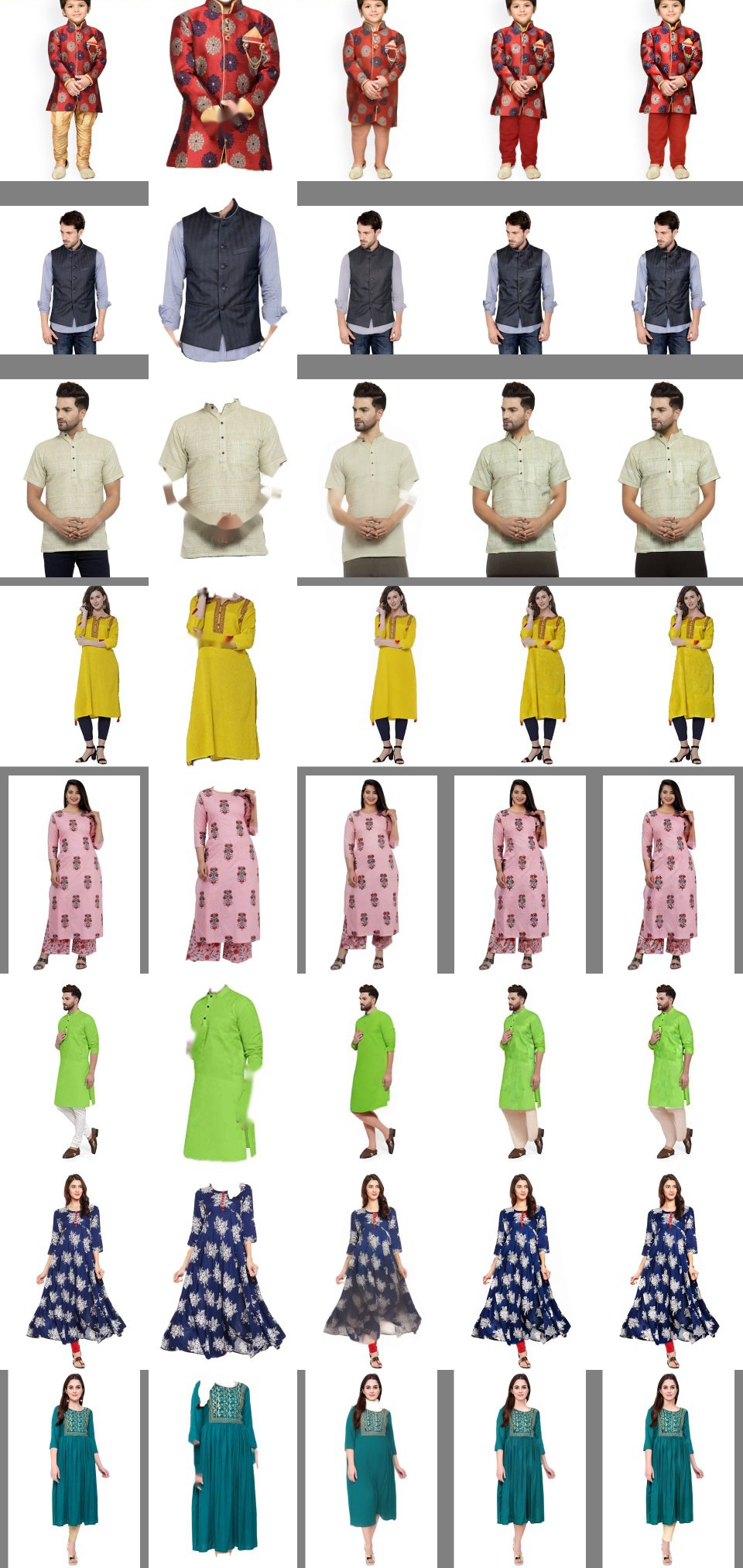}
\captionof{figure}{All eight IndoFashion held-out examples shown in the evaluation grid (person, card, base, Run A, Run B). Row 2: the Nehru jacket keeps the shirt sleeves beneath it in all checkpoints because it uses the \uppermask{} mask. Row 3: a short kurta, also \uppermask{}.}
\label{fig:grid_indo}
\end{center}

\begin{center}
\includegraphics[width=\linewidth]{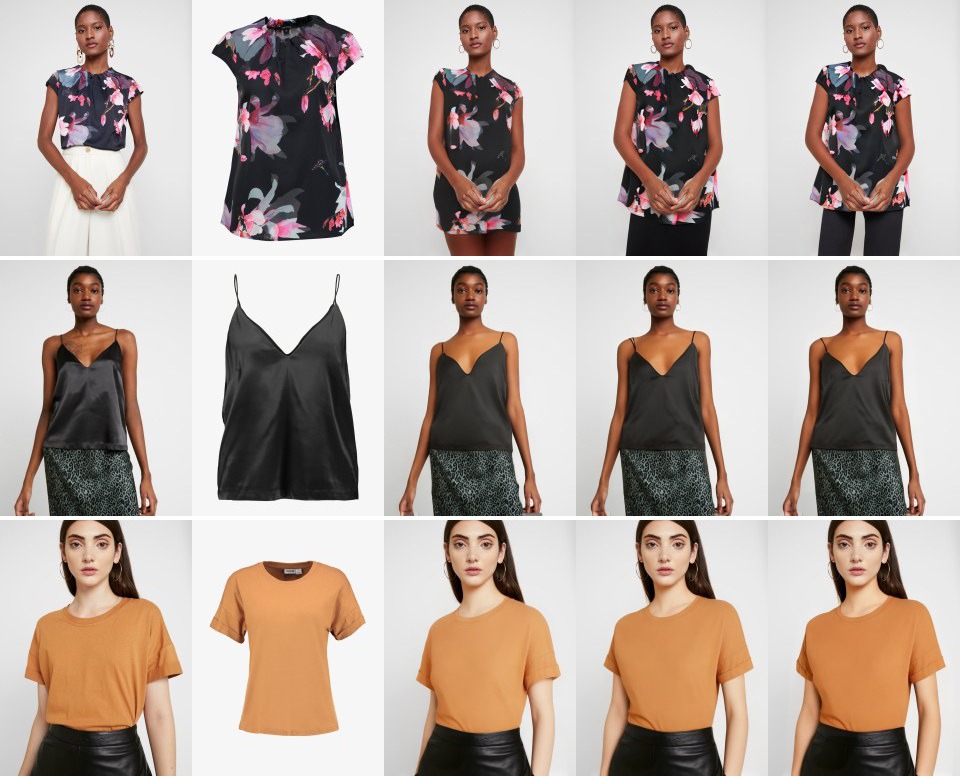}
\captionof{figure}{VITON-HD held-out examples (Western tops, flat-lay cards, \uppermask{} mask). The three checkpoints are visually indistinguishable, matching the flat metrics in Table~\ref{tab:main}.}
\label{fig:grid_vhd}
\end{center}

\begin{center}
\includegraphics[width=\linewidth]{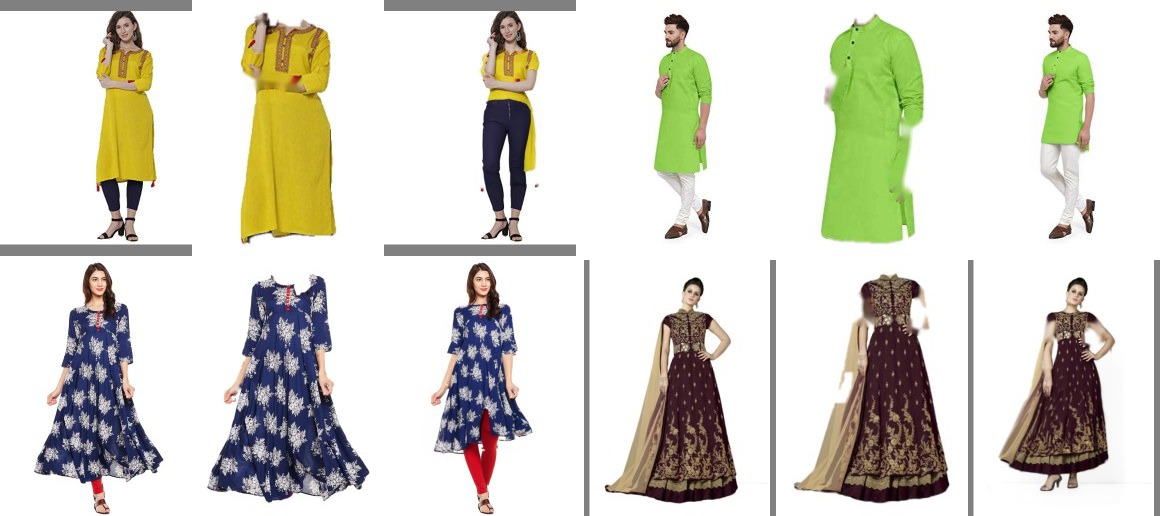}
\captionof{figure}{FASHN VTON 1.5 (masked mode) on held-out ethnic-wear pairs (person, card, FASHN). The knee-length kurtas and the anarkali are cut to hip length and trousers or leggings are invented below: the hem is lost instead of the trousers.}
\label{fig:fashn}
\end{center}

\end{document}